\documentclass[preprint,12pt]{elsarticle}

\usepackage{amssymb}
\usepackage{amsmath}
\usepackage{graphicx}%
\usepackage{multirow}%
\usepackage{amsmath,amssymb,amsfonts}%
\usepackage{amsthm}%
\usepackage{mathrsfs}%
\usepackage[title]{appendix}%
\usepackage{xcolor}%
\usepackage{textcomp}%
\usepackage{manyfoot}%
\usepackage{booktabs}%
\usepackage{algorithm}%
\usepackage{algorithmicx}%
\usepackage{algpseudocode}%
\usepackage{listings}%
\usepackage{pifont}%
\usepackage{hyperref}%

\journal{}

\begin{document}

\begin{frontmatter}

\title{PMT-MAE: Dual-Branch Self-Supervised Learning with Distillation for Efficient Point Cloud Classification}

\author[label1]{Qiang Zheng}
\author[label1]{Chao Zhang}
\author[label1]{Jian Sun \corref{cor1}}

\begin{abstract}
Advances in self-supervised learning are essential for enhancing feature extraction and understanding in point cloud processing. This paper introduces PMT-MAE (Point MLP-Transformer Masked Autoencoder), a novel self-supervised learning framework for point cloud classification. PMT-MAE features a dual-branch architecture that integrates Transformer and MLP components to capture rich features. The Transformer branch leverages global self-attention for intricate feature interactions, while the parallel MLP branch processes tokens through shared fully connected layers, offering a complementary feature transformation pathway. A fusion mechanism then combines these features, enhancing the model's capacity to learn comprehensive 3D representations. Guided by the sophisticated teacher model Point-M2AE, PMT-MAE employs a distillation strategy that includes feature distillation during pre-training and logit distillation during fine-tuning, ensuring effective knowledge transfer. On the ModelNet40 classification task, achieving an accuracy of 93.6\% without employing voting strategy, PMT-MAE surpasses the baseline Point-MAE (93.2\%) and the teacher Point-M2AE (93.4\%), underscoring its ability to learn discriminative 3D point cloud representations. Additionally, this framework demonstrates high efficiency, requiring only 40 epochs for both pre-training and fine-tuning. PMT-MAE's effectiveness and efficiency render it well-suited for scenarios with limited computational resources, positioning it as a promising solution for practical point cloud analysis.
\end{abstract}



\begin{keyword}
Point Cloud \sep Classification \sep Distillation \sep Pre-training \sep Masked Autoencoder
\end{keyword}

\end{frontmatter}

\section{Introduction}
Self-supervised learning is pivotal in artificial intelligence, especially in natural language processing~\cite{GPT2020, BERT2019}, computer vision~\cite{SimSiam2021, MAE2022}, and cross-modal learning~\cite{ALIGN2021, CLIP2021}. This approach excels at extracting robust feature representations from unlabeled data, which can be refined through fine-tuning on specific tasks. The innovation of masked autoencoders (MAE)~\cite{MAE2022} has introduced an asymmetric encoder-decoder architecture that reconstructs occluded regions in 2D images, thereby learning sophisticated latent features. Recently, point cloud analysis has shifted towards self-supervised learning with MAE, focusing on reconstructing masked local patches within point clouds to learn latent 3D representations~\cite{Point-MAE2022, Point-M2AE2022, ACT2023, RECON2023, I2PMAE2023}. This strategy enriches the feature space and paves the way for more efficient analysis techniques for various point cloud processing tasks.

However, integrating effective information in point cloud to enhance feature representation with MAE remains challenging. MAE-based methods can be divided into single-modal and cross-modal approaches, each with limitations. Cross-modal MAE models~\cite{Joint-MAE2023, RECON2023, I2PMAE2023}, though effective, require auxiliary knowledge from other modalities, leading to computational burdens and a reliance on large paired datasets, which may not always be available. Single-modal MAE models~\cite{Point-MAE2022, Point-M2AE2022}, utilizing only point cloud data, are more practical due to their simplicity and efficiency. Models like Point-MAE~\cite{Point-MAE2022} employ global random masking and global self-attention mechanisms, excelling at capturing robust global features but lacking in fine-grained local details. Point-M2AE~\cite{Point-M2AE2022} addresses this by incorporating hierarchical structures and local attention, but its complex multi-scale masking strategy generates numerous intermediate results, demanding substantial storage and computational resources. Point-FEMAE~\cite{PointFEMAE2024} introduces both global and local attention branches, with the local branch enhanced by an Edge-Conv module~\cite{DGCNN2019} for local feature extraction. All the aforementioned designs contribute to increased model complexity and computational load. The challenge is how single-modal MAE models can efficiently integrate global and local information to strengthen feature representation while maintaining computational efficiency. Additionally, the pre-training task of reconstructing a large proportion of masked, unlabeled patches often requires many iterations for convergence, further increasing computational resource consumption. Addressing these challenges is crucial for advancing single-modal MAE models in point cloud analysis.

To enhance the feature representation capabilities of MAE model, we introduce a parallel Multilayer Perceptron (MLP) branch within the conventional MAE architecture, denoted as PMT-MAE. This branch processes each token independently through a shared fully-connected layer, providing a distinct feature transformation pathway. The outputs from the MLP branch are then directly concatenated with the tokens from the Transformer branch, followed by a fusion process that integrates these complementary features. This innovative approach offers several advantages: it enhances the model's ability to capture diverse patterns and structural information from point cloud data, accelerates convergence during training by providing a straightforward and efficient feature transformation, and enriches the model's representational power, potentially leading to improved performance across various downstream tasks. This approach maintains computational efficiency, ensuring that the MLP branch's benefits do not come at the cost of excessive resource utilization. Furthermore, the dual-branch structure operates on a global scale due to the masking operation's disruption of fine-grained features.

Additionally, inspired by knowledge distillation techniques, particularly in BERT model compression~\cite{TinyBERT2020}, we have integrated a two-stage distillation strategy within our PMT-MAE framework. During the pre-training phase, the model undergoes general distillation, akin to acquiring foundational knowledge. This is followed by task-specific distillation during fine-tuning, where the model refines its abilities for targeted tasks. This two-stage distillation facilitates efficient knowledge transfer from a well-trained teacher model, Point-M2AE~\cite{Point-M2AE2022}, which has a profound understanding of the data. This method ensures that PMT-MAE explores global features autonomously and receives fine-grained details from the teacher model, enhancing its feature representation capabilities. The two-stage process allows PMT-MAE to capture data nuances at both macro and micro levels, enriching its understanding and improving task performance. Moreover, this approach leads to more efficient convergence and conserves computational resources under the guidance of the teacher model. By distilling knowledge during both pre-training and fine-tuning, PMT-MAE assimilates a rich understanding of the data's underlying structure and task-specific features more rapidly, enhancing overall training efficiency.

The contributions of this study are as follows:
\begin{itemize}
\item The design of the PMT-MAE framework, featuring a dual-branch structure that includes both a Transformer and an MLP branch, is engineered to extract comprehensive features by leveraging parallel processing pathways.
\item The incorporation of a two-stage distillation strategy allows PMT-MAE to assimilate knowledge from the teacher model, Point-M2AE~\cite{Point-M2AE2022}, during pre-training and fine-tuning, enriching feature representation and accelerating convergence.
\item PMT-MAE outperforms both the baseline and teacher models on point cloud classification tasks, demonstrating faster convergence and reduced computational costs.
\end{itemize}

\section{Related Works}
\subsection{Supervised Learning for Point Cloud Analysis}
The field of point cloud analysis has been significantly advanced by a variety of supervised learning methods that have been developed to address the unique challenges of processing 3D data. Among these, the PointNet~\cite{PointNet2017} architecture stands out for its ability to directly consume point clouds and demonstrated impressive performance on tasks such as 3D classification and segmentation. This work was further extended to capture more fine-grained patterns through a hierarchical neural network known as PointNet++~\cite{PointNetplus2017}, which applies PointNet~\cite{PointNet2017} recursively on nested partitions of the point set.

In parallel, the adaptation of convolutional neural networks (CNNs) to the irregular and unordered nature of point clouds has been a significant endeavor. Li et al. proposed PointCNN~\cite{PointCNN2018}, which learns an $\chi$-transformation to address the issues of spatially-local correlation and point ordering, thus generalizing CNNs to feature learning from point clouds. Wang et al. introduced a novel approach with deep parametric continuous convolutional neural networks (PCNN)~\cite{Deep2021}, which utilize parameterized kernel functions over non-grid structured data, showing significant improvements in point cloud segmentation.

The integration of graph neural networks (GNNs) with point cloud analysis has also emerged as a powerful approach. Wang et al. presented a dynamic graph CNN (DGCNN)~\cite{DGCNN2019} that dynamically updates the graph structure based on feature space, allowing for the capture of both local and global geometric properties. Zhou et al. introduced Adaptive Graph Convolution (AdaptConv)~\cite{Adaptive2021}, which generates adaptive kernels for points according to their dynamically learned features, enhancing the flexibility and precision of point cloud convolutions.

Furthermore, the transformative impact of Transformer architectures on point cloud processing has been recognized with the introduction of the Point Transformer~\cite{point-trans2021}. This model leverages self-attention mechanisms to process point clouds [6]. The subsequent Point Transformer V2~\cite{point-transV2-2022} refines the approach with grouped vector attention and partition-based pooling, showcasing further advancements in efficiency and effectiveness.

\subsection{Self-Supervised Learning for Point Cloud Analysis}
Self-supervised learning (SSL) has emerged as a powerful paradigm for point cloud analysis, offering a means to leverage large amounts of unlabeled data by creating surrogate tasks for the network to solve. SSL methods can be broadly categorized into discriminative~\cite{SimCLR2020, 3DVLP2024} and generative approaches~\cite{MAE2022, data2vec2022}. In the realm of 3D point cloud analysis, discriminative self-supervised learning has been explored through methods like PointContrast~\cite{PointContrast2020} and CrossPoint~\cite{CrossPoint2022}. PointContrast~\cite{PointContrast2020} leverages contrastive learning to pre-train for 3D point cloud understanding. Similarly, CrossPoint~\cite{CrossPoint2022} employs a cross-modal contrastive learning strategy to learn transferable 3D point cloud representations by relating them to 2D images. Generative methods, on the other hand, involve learning to generate data as a means to understand its structure. A prominent example in the image domain is the Masked Autoencoder (MAE)~\cite{MAE2022}, which masks parts of the input and learns to reconstruct them, fostering a deep understanding of the data distribution. This concept has been adapted to point cloud analysis, leading to the development of MAE-based methods tailored for the unique challenges posed by 3D data. The subsequent section delves into the advancements and applications of MAE-based point cloud analysis.

MAE-based methods in point cloud analysis can be categorized into single-modal and cross-modal approaches. Single-modal methods concentrate solely on 3D point clouds, exemplified by CP-Net~\cite{CP-Net2024}, Point-BERT~\cite{Point-BERT2022}, Point-MAE~\cite{Point-MAE2022}, and Point-M2AE~\cite{Point-M2AE2022}. CP-Net~\cite{CP-Net2024} introduces a contour-perturbed augmentation module to enhance semantic learning by perturbing point cloud contours while preserving content. Point-BERT~\cite{Point-BERT2022} employs a Masked Point Modeling task, utilizing discrete Variational Autoencoders for local feature encapsulation and Transformer pre-training. Point-MAE~\cite{Point-MAE2022} applies masked autoencoding to learn latent features from randomly masked 3D points, while Point-M2AE~\cite{Point-M2AE2022} employs multi-scale architectures to capture hierarchical geometrical semantics. In contrast, cross-modal approaches such as Joint-MAE~\cite{Joint-MAE2023}, ACT~\cite{ACT2023}, I2P-MAE~\cite{I2PMAE2023}, and Recon~\cite{RECON2023} integrate diverse modalities to enhance point cloud analysis. Joint-MAE~\cite{Joint-MAE2023} innovatively combines 2D-3D joint Masked Autoencoders to exploit semantic and geometric correlations between images and point clouds. ACT~\cite{ACT2023} leverages pre-trained 2D Transformers to guide 3D representation learning through self-supervised autoencoding. I2P-MAE~\cite{I2PMAE2023} bridges the gap between limited 3D datasets and abundant 2D data by utilizing multi-view features for point cloud reconstruction. Recon~\cite{RECON2023} unifies contrastive and generative models, employing an ensemble distillation approach to optimize performance across diverse datasets. Given the considerations of computational efficiency and data availability, this study adopts a single-modal scheme combined with a distillation strategy to optimize point cloud analysis.

\section{Methodology}
\subsection{MAE for Computer Vision}
The burgeoning field of self-supervised learning has seen the rise of Masked Autoencoders (MAEs)~\cite{MAE2022}, which have garnered significant attention for their efficacy in computer vision tasks. MAEs differ from traditional supervised learning approaches by learning from unlabeled data, thereby avoiding the labor-intensive process of manual annotation. This paradigm shift is particularly notable in feature representation learning, where MAEs demonstrate an exceptional ability to discern and reconstruct visual information.

The Masked Autoencoder (MAE) is based on an asymmetric encoder-decoder structure designed to efficiently handle masked input images. At its core, the encoder processes only the visible portions of the image, strategically ignoring the masked regions and producing a condensed latent representation that captures the essence of the exposed visual data. This architectural asymmetry is not just a design choice but a strategic move that enables the model to learn and generalize effectively from incomplete data. The MAE operates by initially applying random masking to the input image, creating an environment where the model must deal with incomplete information. Despite this challenge, the encoder encodes the visible patches into a latent space, while the decoder, equipped with this latent representation and mask tokens, is tasked with reconstructing the complete image, including the initially occluded regions. This reconstruction task is central to the model's learning process, compelling the network to develop robust and comprehensive feature representations.

\begin{figure}[ht]
    \centering
    \includegraphics[width=0.7\linewidth]{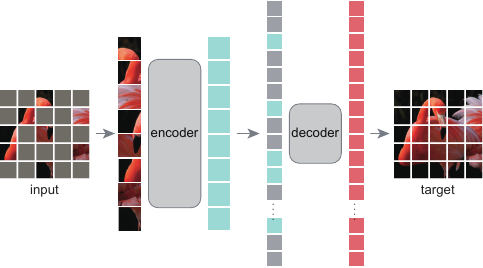}
    \caption{MAE Framework for Visual Learning~\cite{Point-MAE2022}.}
    \label{fig-PMT-MAE-Image-MAE}
\end{figure}

Fig.~\ref{fig-PMT-MAE-Image-MAE} provides a visual representation of the MAE workflow during the pre-training phase, illustrating the process where a substantial subset of image patches is masked out. The encoder processes only the visible patches, and after encoding, mask tokens are introduced. The lightweight decoder then reconstructs the original image in pixel space from the full set of encoded patches and mask tokens. This visual aid underscores the efficiency of training large models with MAEs and the acceleration of the learning process due to the reduced computational burden on the encoder. The MAE's mechanism is straightforward yet effective, and its ability to handle incomplete data through the strategic use of masking and reconstruction makes it a powerful tool for self-supervised learning in the field of computer vision.

MAEs have applications in various computer vision tasks, including object recognition, scene understanding, and image generation. Pre-training on large datasets with masked images allows MAEs to learn generalizable features transferable to multiple downstream tasks. This transferability highlights the versatility of MAEs, which can serve as feature extractors or be fine-tuned on task-specific datasets, enhancing performance across diverse vision applications.

In summary, MAEs offer a compelling approach to unsupervised learning in computer vision, characterized by their ability to learn from incomplete data and generalize to a wide range of tasks. The architectural nuances, operational mechanisms, and applications of MAEs, as elucidated with the aid of Fig.~\ref{fig-PMT-MAE-Image-MAE}, highlight their promise as a cornerstone for advancing the field of computer vision.

\subsection{Architecture Comparison and Selection}
\begin{figure}[ht]
    \centering
    \includegraphics[width=1.0\linewidth]{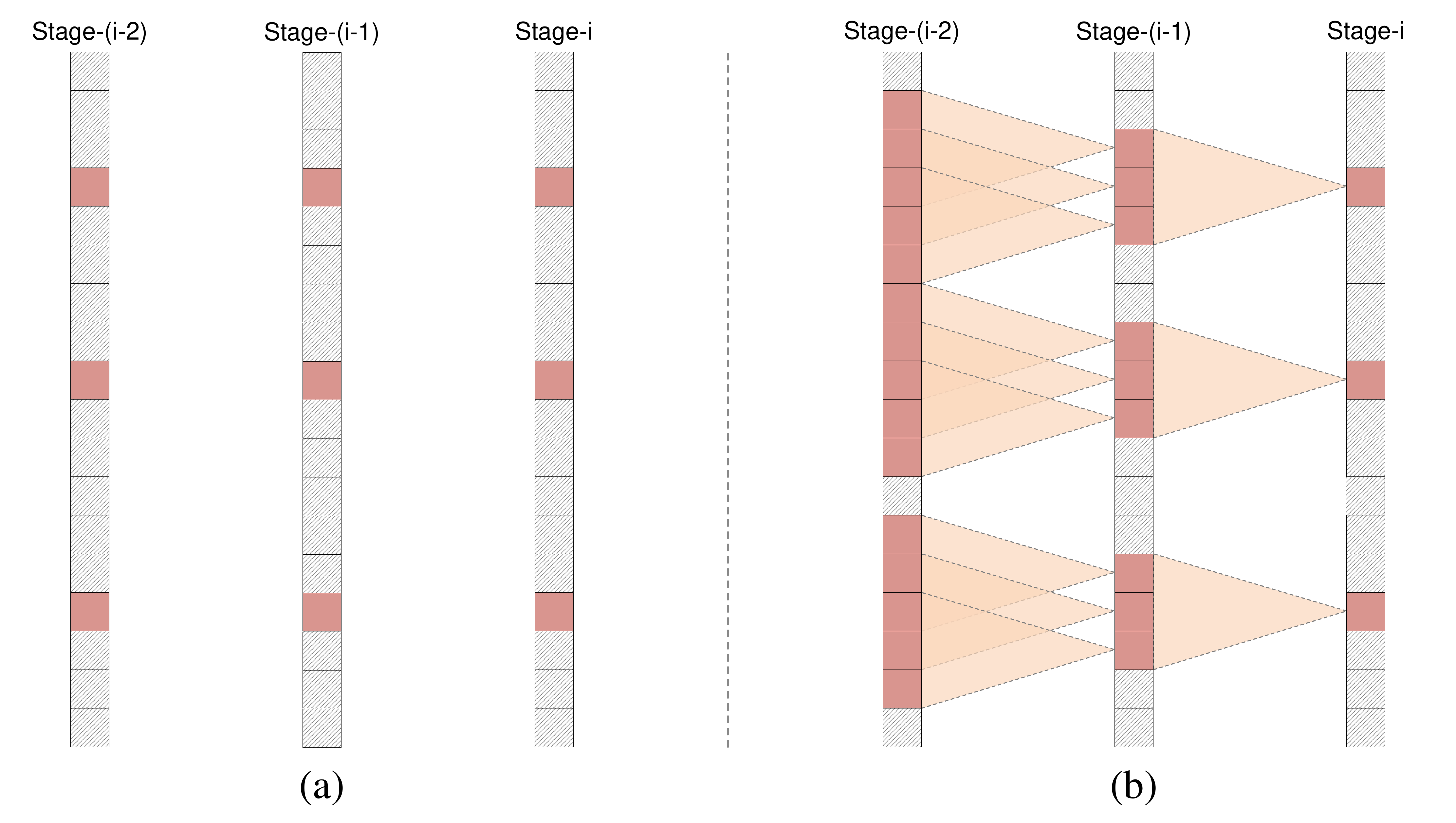}
    \caption{Masking strategies for Point-MAE~\cite{Point-MAE2022} (a) and Point-M2AE~\cite{Point-M2AE2022} (b). Red and shaded blocks represent visible and masked tokens, respectively. Point-MAE~\cite{Point-MAE2022} employs a non-pyramidal structure with consistent masking across stages, while Point-M2AE~\cite{Point-M2AE2022}'s pyramidal structure necessitates complex masking operations in reverse stage order. This leads to inconsistent visible token counts, increased computational and memory overhead, and a diminishing effective masking rate with shallower stages, hindering model scalability.}
    \label{fig-PMT-MAE-Mask-Mechanism}
\end{figure}

In the domain of self-supervised learning for point clouds, the architectural decisions of Point-MAE~\cite{Point-MAE2022} and Point-M2AE~\cite{Point-M2AE2022} present distinct approaches to feature representation and extraction, as shown in Fig.~\ref{fig-PMT-MAE-Mask-Mechanism}. Point-MAE~\cite{Point-MAE2022} employs a non-pyramidal, or straight-through, architecture, characterized by a direct and efficient flow of information from input to output without the complexities of hierarchical processing. This streamlined structure offers several advantages, including simplified modeling, strong scalability, and a focus on global feature representation. However, the non-pyramidal design may not capture local fine-grained details as effectively as pyramidal structures, which is a limitation when dealing with intricate point cloud features.

Conversely, Point-M2AE~\cite{Point-M2AE2022} adopts a pyramidal architecture, which integrates multi-scale feature extraction through a series of downsampling and upsampling operations. This approach adeptly captures both local and global features, providing a comprehensive representation of point cloud data. The pyramidal structure aligns with the inductive bias that leverages spatial hierarchies to enhance model performance on complex tasks. However, Point-M2AE~\cite{Point-M2AE2022}'s pyramidal structure introduces significant complexity, particularly in the masking mechanism. The process of propagating masks from deeper to shallower stages necessitates intricate operations that ensure consistent visibility across scales. This complexity not only escalates computational demands but also poses challenges in scaling the model to deeper architectures. The mask propagation strategy can lead to an expansion of visible points in shallower stages, undermining the effectiveness of the masking strategy and potentially diluting the model's ability to learn from incomplete data. Moreover, the complex mask handling in Point-M2AE~\cite{Point-M2AE2022} can result in increased memory usage and slower inference times, which are critical considerations for real-world applications, especially on resource-constrained platforms.

Based on this comparative analysis, the current study opts for the non-pyramidal structure of Point-MAE~\cite{Point-MAE2022}. This choice is driven by the desire for a modeling framework that offers simplicity, ease of training, and the capacity to generalize effectively across a variety of point cloud tasks. While acknowledging the limitations in capturing local details, the non-pyramidal structure's merits in terms of scalability, computational efficiency, and preservation of global context outweigh these drawbacks for the scope of this research.

To address the non-pyramidal structure's limitation in capturing local fine-grained features, this study employs Point-M2AE~\cite{Point-M2AE2022} as a teacher model in a knowledge distillation framework. By transferring the knowledge encoded in the teacher model to the student model, the study integrates the advantages of both architectures. The student model, through the distillation process, acquires the ability to perceive local details that are essential for nuanced feature representation while maintaining the efficiency and global context understanding of the non-pyramidal structure.

In conclusion, the architectural choice in this study is underpinned by a comprehensive consideration of the trade-offs between complexity and efficiency, and the need for a model that balances the requirements of point cloud processing tasks with the practical constraints of deployment and computation. The non-pyramidal structure of Point-MAE~\cite{Point-MAE2022}, complemented by knowledge distillation from the pyramidal Point-M2AE~\cite{Point-M2AE2022}, provides a balanced approach that harnesses the strengths of both architectural paradigms.

\subsection{Dual-Branch Structure}
\begin{figure}[htbp]
    \centering
    \includegraphics[width=1.0\linewidth]{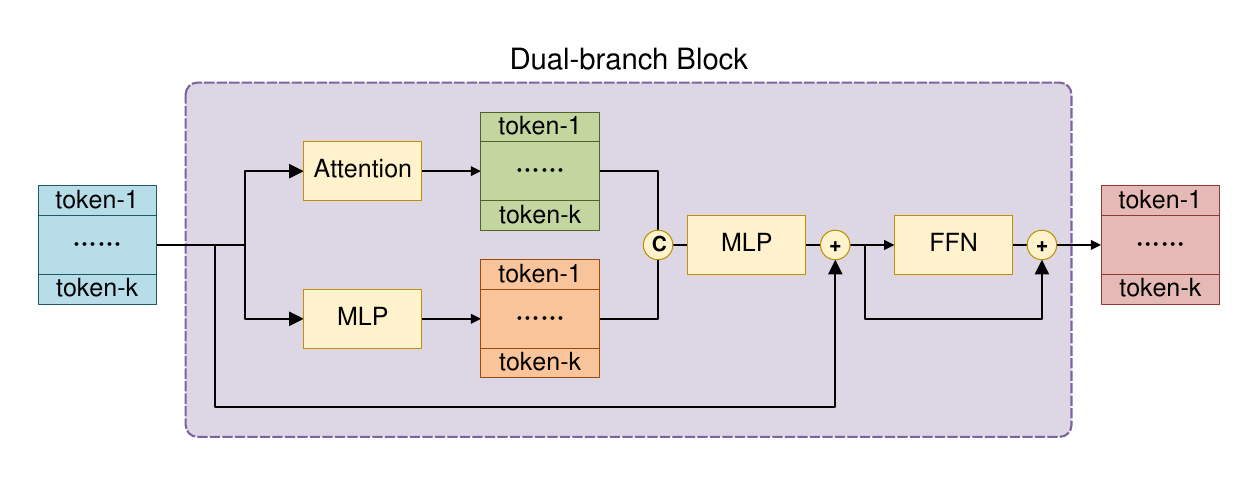}
    \caption{Illustration of the dual-branch block.}
    \label{fig-PMT-MAE-Dual-Branch}
\end{figure}

The dual-branch structure presented in this research integrates MLPs with Transformers, as illustrated in Fig.~\ref{fig-PMT-MAE-Dual-Branch}. This architecture fuses the strengths of both MLPs and Attention mechanisms, facilitating a more efficient and effective approach to global feature extraction in point cloud analysis. By parallelizing the MLP branch with the Attention branch, and subsequently fusing their outputs before feeding into the Feed-Forward Network (FFN) module, this structure overcomes the individual limitations of each approach. In contrast to previous work~\cite{PointMT2024}, the dual-branch architecture presented here operates on a global scale for both branches, due to the masking mechanism's destruction of local information.

Traditional point cloud analysis methods often rely solely on either MLPs or Transformers, each with inherent drawbacks. MLP-based approaches, while simple and fast to converge, often neglect the inter-point correlations, leading to suboptimal feature characterization. Conversely, Transformer-based methods, though adept at aggregating features through weighted attention, suffer from complex modeling and slow convergence. The proposed dual-branch structure addresses these limitations by combining the rapid convergence of MLPs with the nuanced feature aggregation capabilities of Transformers, thereby enhancing overall performance and efficiency.

The dual-branch structure operates by processing the input data through two parallel branches: an MLP branch and an Attention branch. The MLP branch performs independent transformations on individual points via shared fully connected layers, ensuring quick and efficient processing. Simultaneously, the Attention branch assigns weights to points based on their importance, facilitating refined feature aggregation. The outputs from both branches are concatenated and then passed through an FFN module, which integrates the diverse feature representations and produces the final output. This approach ensures a coherent feature transmission channel and accelerates model convergence.

The architecture is designed to leverage the strengths of each branch. The MLP branch processes each point independently using shared fully connected layers, which simplifies the model and speeds up convergence. Despite its simplicity, it lacks the ability to capture inter-point relationships. On the other hand, the Attention branch uses the Attention mechanism to weigh the importance of each point, allowing for sophisticated feature aggregation. It addresses the MLP branch's limitation by considering the correlations between points. After processing in both branches, their outputs are concatenated and then processed in the FFN module, resulting in a robust and comprehensive feature representation.

The proposed dual-branch structure offers several advantages over traditional methods. By combining MLP and Attention mechanisms, the model can capture both individual point features and their inter-relations. Additionally, the MLP branch ensures fast convergence, while the Attention branch provides nuanced feature aggregation, striking a balance between complexity and efficiency. The parallel branches allow for independent optimization of feature representations, enhancing the model's adaptability to various point cloud tasks.

\subsection{Two-stage Distillation Framework}
\begin{figure}[p]
    \centering
    \includegraphics[width=1.0\linewidth]{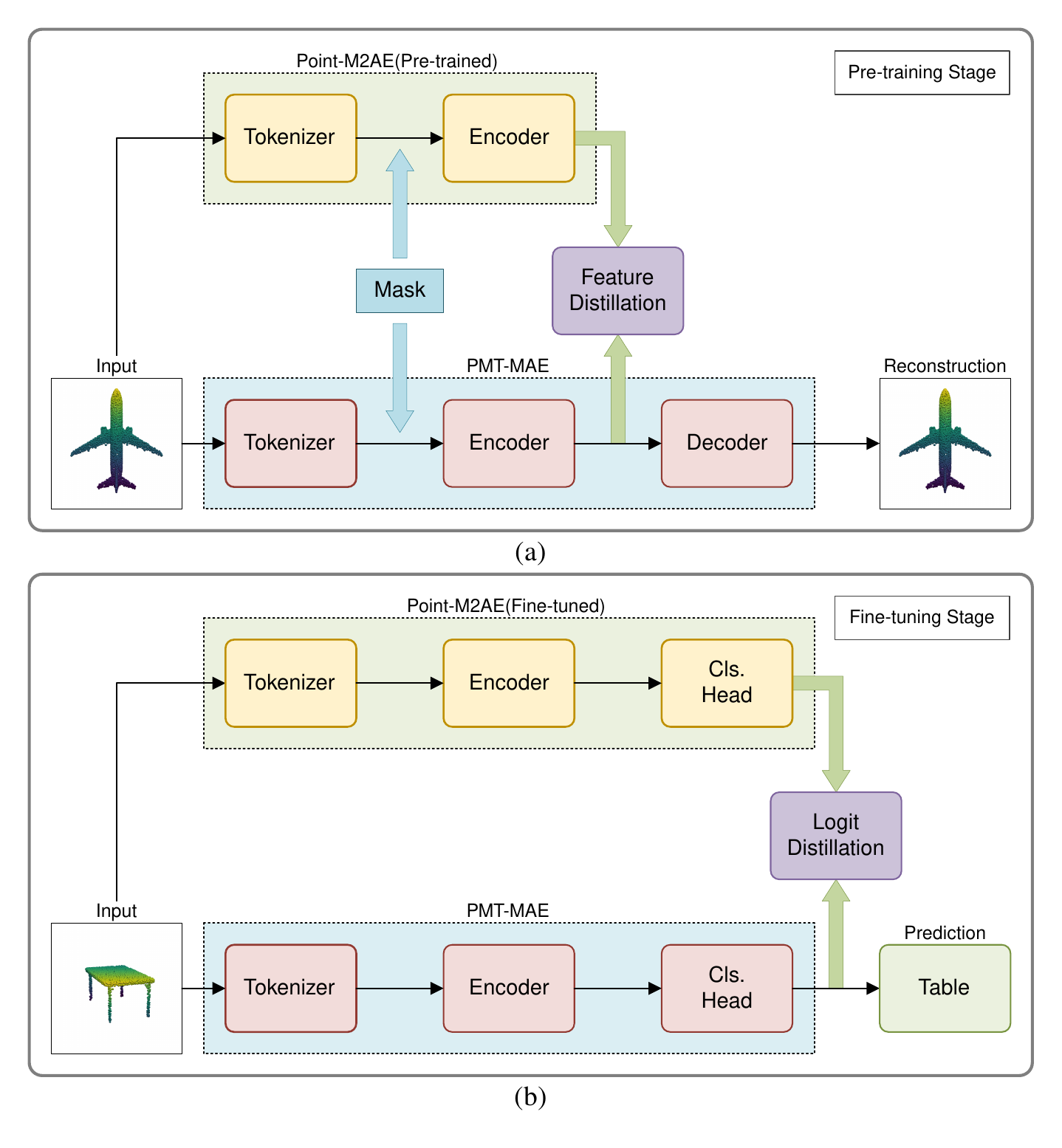}
    \caption{The two-stage distillation framework used in PMT-MAE includes distinct processes for the pre-training and fine-tuning phases, leveraging the strengths of the Point-M2AE~\cite{Point-M2AE2022} model. (a) In the pre-training phase, the framework employs the pre-trained model of Point-M2AE~\cite{Point-M2AE2022} for feature distillation. Here, the encoder of PMT-MAE is guided to replicate the output features of the Point-M2AE~\cite{Point-M2AE2022} encoder, ensuring that the student model effectively captures meaningful feature representations. (b) During the fine-tuning phase, the framework addresses the point cloud classification problem by utilizing the fine-tuned model of Point-M2AE~\cite{Point-M2AE2022}. In this phase, the output of PMT-MAE undergoes logit distillation, focusing on transferring the classification prediction knowledge from the teacher model to the student model. This two-stage approach enhances the classification accuracy and robustness of PMT-MAE, providing a comprehensive strategy for effective knowledge transfer.}
    \label{fig-PMT-MAE-Distillation-Framework}
\end{figure}

Inspired by TinyBERT~\cite{TinyBERT2020}'s innovative knowledge distillation approach, this study employs a two-stage distillation framework to enhance the performance of the proposed PMT-MAE model for point cloud analysis. While TinyBERT~\cite{TinyBERT2020} focuses on BERT~\cite{BERT2019} distillation, our adaptation applies this framework to transfer knowledge from the Point-M2AE~\cite{Point-M2AE2022} teacher model, effectively addressing the limitations of the PMT-MAE network in extracting local fine-grained information. The two-stage distillation strategy in this study features a simpler design compared to TinyBERT~\cite{TinyBERT2020}, focusing solely on the output features of the encoder without involving intermediate features across multiple layers, shown in Fig.~\ref{fig-PMT-MAE-Distillation-Framework}. This simplification arises due to the significant structural differences between the teacher and student models. Specifically, the number of points, masking situations, and feature alignment within the intermediate layers of the encoder vary greatly between the two models, making alignment operations impractical.

The two-stage distillation framework operates in two distinct stages: pre-training (feature distillation) and fine-tuning (logit distillation). During the pre-training stage, the PMT-MAE model is trained on the ShapeNet-55 dataset to simulate the output features of the encoder of the Point-M2AE model. This involves the student model replicating the intermediate representations of the teacher model. To ensure consistency, the student model also receives the mask flags from the teacher encoder at the output layer, along with the teacher’s output features. This approach focuses on distilling the encoder features, which are critical for downstream tasks, while avoiding the complexities and potential overfitting issues associated with comparing the reconstruction results of the decoder. During the pre-training stage, the total loss is a combination of the feature distillation loss and the reconstruction loss of the point cloud patch. The weight of the feature distillation loss is set to \(\alpha\).

\begin{equation}
\mathcal{L}_{\text{feat}} = \sum_{i=1}^{K} \left\| \mathbf{F}_i^{\text{tea}} - \mathbf{F}_i^{\text{stu}} \right\|_2^2
\label{eq-PMT-MAE-feat-distil}
\end{equation}
In Eq.~\ref{eq-PMT-MAE-feat-distil}, \(\mathcal{L}_{\text{feat}}\) represents the feature distillation loss, where \(\mathbf{F}_i^{\text{tea}}\) and \(\mathbf{F}_i^{\text{stu}}\) denote the output features of the \(i\)-th token from the teacher and student models, respectively. \(K\) is the total number of tokens.

\begin{equation}
\mathcal{L}_{\text{recon}} = \frac{1}{M_{re}} \sum_{{p_{re} \in \mathbf{P_{re}}}}^{} \min_{p_{gt} \in \mathbf{P_{gt}}} \left\| p_{re} - p_{gt} \right\|_2^2 + \frac{1}{M_{gt}} \sum_{{p_{gt} \in \mathbf{P_{gt}}}}^{} \min_{p_{re} \in \mathbf{P_{re}}} \left\| p_{gt} - p_{re} \right\|_2^2
\label{eq-PMT-MAE-recon}
\end{equation}
In Eq.~\ref{eq-PMT-MAE-recon}, \(\mathcal{L}_{\text{recon}}\) represents the reconstruction loss of the point cloud patch, calculated using the \(L_2\) Chamfer Distance. Moreover, \(p_{gt}\) denotes a point in the original point cloud \(\mathbf{P_{gt}}\), and \(p_{re}\) denotes a point in the reconstructed point cloud \(\mathbf{P_{re}}\). $M_{gt}$ and $M_{re}$ represent the number of points contained in the original and reconstructed point cloud patches, respectively.

\begin{equation}
\mathcal{L}_{\text{total}} = \alpha \mathcal{L}_{\text{feat}} + \mathcal{L}_{\text{recon}}
\label{eq-PMT-MAE-pretrain}
\end{equation}
In Eq.~\ref{eq-PMT-MAE-pretrain}, \(\mathcal{L}_{\text{total}}\) represents the total loss for the pre-training stage, which is a weighted sum of the feature distillation loss and the reconstruction loss, with \(\alpha\) being the weight of the feature distillation loss.

In the fine-tuning stage, the distillation process shifts to logit distillation, where the PMT-MAE model learns to replicate the classification predictions (logits) of the Point-M2AE~\cite{Point-M2AE2022} teacher model. This stage further refines the model’s feature representations and ensures optimal adaptation to specific point cloud tasks, enhancing the model's overall performance. During the fine-tuning stage, the total loss is a combination of the logit distillation loss and the cross-entropy loss for classification. The weight of the logit distillation loss is set to \(\beta\).

\begin{equation}
\mathcal{L}_{\text{logit}} = \frac{1}{N} \sum_{i}^{N} {T^2} D_{\text{KL}}(\mathbf{z}_i^{\text{tea}} || \mathbf{z}_i^{\text{stu}})
\label{eq-PMT-MAE-logit-distil}
\end{equation}
In Eq.~\ref{eq-PMT-MAE-logit-distil}, \(\mathcal{L}_{\text{logit}}\) represents the logit distillation loss, where \(D_{\text{KL}}\) denotes the Kullback-Leibler (KL) divergence between the logits of the teacher model \(\mathbf{z}_i^{\text{tea}}\) and the student model \(\mathbf{z}_i^{\text{stu}}\) for the \(i\)-th sample. The temperature parameter \(T\) is used to soften the probability distributions, with \(T^2\) scaling the KL divergence. \(N\) is the total number of samples in the batch.

\begin{equation}
\mathcal{L}_{\text{CE}} = - \frac{1}{N} \sum_{i=1}^{N} y_i \log \hat{y}_i
\label{eq-PMT-MAE-CE}
\end{equation}
In Eq.~\ref{eq-PMT-MAE-CE}, \(\mathcal{L}_{\text{CE}}\) represents the cross-entropy loss for classification, where \(y_i\) is the ground truth label and \(\hat{y}_i\) is the predicted probability for the \(i\)-th sample. The loss is averaged over the total number of samples \(N\) in the batch.

\begin{equation}
\mathcal{L}_{\text{total}} = \beta \mathcal{L}_{\text{logit}} + \mathcal{L}_{\text{CE}}
\label{eq-PMT-MAE-finetune}
\end{equation}
In Eq.~\ref{eq-PMT-MAE-finetune}, \(\mathcal{L}_{\text{total}}\) represents the total loss for the fine-tuning stage, which is a weighted sum of the logit distillation loss and the cross-entropy loss, with \(\beta\) being the weight of the logit distillation loss.

Several key designs contribute to the effectiveness of the two-stage distillation framework in this study. During the pre-training stage, the student model mimics the output features of the teacher model's encoder, focusing on feature distillation. The student model receives the mask flags of the teacher model, ensuring consistency in the mask points between both models. This approach avoids the complexities of reconstructing the teacher model's decoder output, which could lead to overfitting. Instead, it emphasizes the importance of the teacher model's encoder features for downstream tasks. During the fine-tuning stage, logit distillation occurs, where the student model distills the classification prediction logits from the teacher model, further enhancing its performance. This two-stage process ensures that the student model gains a robust understanding of both feature representation and classification accuracy. Given the hierarchical nature of the Point-M2AE~\cite{Point-M2AE2022} model, the two-stage distillation framework adeptly compensates for the PMT-MAE model's non-pyramidal architecture, enriching its global feature representation with local fine-grained information.

The two-stage distillation framework offers significant advantages. By transferring knowledge from the Point-M2AE~\cite{Point-M2AE2022} teacher model, the PMT-MAE model effectively captures both global and local features, resulting in a more comprehensive and nuanced feature representation crucial for accurate point cloud analysis. Additionally, the two-stage distillation process significantly accelerates the convergence of the PMT-MAE model. The enriched feature representations learned during the distillation stages reduce the number of iterations required for training, making the model more efficient and faster to train.

\section{Experiments}
This section begins by detailing the model structure and parameter settings. It then presents the experimental results for point cloud classification, followed by an analysis of the model's complexity. Subsequently, ablation experiments are conducted to assess the impact of various components of the model. Finally, visualization results are provided to offer further insights into the model's performance.

\subsection{Configuration}
This section focuses on the network structure of PMT-MAE and the parameter settings for both the pre-training and fine-tuning phases. For details on the network structure of the teacher model Point-M2AE, please refer to the literature \cite{Point-M2AE2022}.

\textbf{Network Structure.} PMT-MAE employs mini-PointNet~\cite{PointNet2017} as a tokenizer to convert local point cloud patches into tokens. The dual-branch block serves as the fundamental unit in both the encoder and decoder. This study explores two versions of PMT-MAE, namely PMT-MAE-S (``S'': Small) and PMT-MAE-L (``L'': Large), distinguished by the number of blocks in the encoder, which are 6 and 12, respectively. Both versions have decoders containing 4 blocks. During the pre-training phase, a masking operation is applied to the token input of the encoder, and the decoder aims to reconstruct the patches corresponding to the masked tokens. In the fine-tuning phase, the output features of the PMT-MAE encoder are fed into a classification head for category prediction.

\textbf{Training Parameters.} The pre-trained and fine-tuned models provided by Point-M2AE \cite{Point-M2AE2022} are utilized in this study, serving as the teacher models in the pre-training and fine-tuning stages, respectively. The pre-trained Point-M2AE model employs a mask rate of 0.8.

In the pre-training phase, the ShapeNet-55 dataset, comprising 52,470 samples across 55 classes, is used without its labels. The input point cloud consists of 2048 points, with 64 patches sampled and a mask rate of 0.7. The input and output channels of the dual-branch blocks are set to 384. A cosine learning rate, decreasing from $1.0 \times 10^{-3}$ to $1.0 \times 10^{-6}$ over 40 epochs, is employed. The weight \(\alpha\) is set to 1.0, and the batch size is 32.

For the fine-tuning stage, the ModelNet40 dataset is used, which includes 12,311 samples across 40 categories, with 9,843 samples in the training set and 2,468 samples in the test set. The input point cloud consists of 1024 points, with 64 patches sampled. A cosine learning rate, decreasing from $1.0 \times 10^{-3}$ to $1.0 \times 10^{-6}$ over 40 epochs, is used. The weight \(\beta\) is set to 0.01, and the temperature \(T\) is set to 3.0. The batch size is 24.

\subsection{Classification Task}
Tab.~\ref{tab-PMT-MAE-cls} presents a comparative analysis of the PMT-MAE model's performance on the ModelNet40 classification task against other existing methods. All results are reported without employing the voting strategy to ensure a fair comparison. These methods are divided into three categories: supervised methods, single-modal self-supervised methods (Self-supervised$\text{[S]}$), and cross-modal self-supervised methods (Self-supervised$\text{[C]}$). PMT-MAE belongs to the single-modal self-supervised category.

\begin{table}[ht]
    \centering
    \setlength{\tabcolsep}{3.0mm}
    \begin{tabular*}{\textwidth}{@{\extracolsep{\fill}}l|ccc}
        \toprule[1pt]
        \textbf{Method}                                 & Type          & Input& Acc.(w/o voting) \\
        \midrule[0.3pt]
        PointNet~\cite{PointNet2017}                    & Supervised    & 1k   & 89.2 \\
        PointNet++(MSG)~\cite{PointNetplus2017}         & Supervised    & 5k   & 91.9 \\
        DGCNN~\cite{DGCNN2019}                          & Supervised    & 1k   & 92.9 \\
        KPConv~\cite{KPConv2019}                        & Supervised    & 6.8k & 92.9 \\
        PT~\cite{point-trans2021}                       & Supervised    & 1k   & 93.7 \\
        PointMLP~\cite{PointMLP2022}                    & Supervised    & 1k   & 94.1 \\
        \midrule[0.3pt]
        CP-Net~\cite{CP-Net2024}                        & Self-supervised$\text{[S]}$  & 1k    & 91.9 \\
        Point-BERT~\cite{Point-BERT2022}                & Self-supervised$\text{[S]}$  & 1k    & 93.2 \\
        Point-MAE~\cite{Point-MAE2022}                  & Self-supervised$\text{[S]}$  & 1k    & 93.2 \\
        Point-M2AE~\cite{Point-M2AE2022}                & Self-supervised$\text{[S]}$  & 1k    & 93.4 \\
        \midrule[0.3pt]
        CrossNet~\cite{CrossNet2024}                    & Self-supervised$\text{[C]}$  & 1k    & 93.4 \\
        Inter-MAE~\cite{Inter-MAE2024}                  & Self-supervised$\text{[C]}$  & 1k    & 93.6 \\
        Joint-MAE~\cite{Joint-MAE2023}                  & Self-supervised$\text{[C]}$  & 1k    & 94.0 \\
        \midrule[0.3pt]
        PMT-MAE-S (Ours)                                & Self-supervised$\text{[S]}$  & 1k    & 93.5 \\
        PMT-MAE-L (Ours)                                & Self-supervised$\text{[S]}$  & 1k    & 93.6 \\
        \bottomrule[1pt]
    \end{tabular*}
    \vspace{5pt}
    \caption{Classification accuracy (\%) on the ModelNet40 dataset. Methods are categorized in the Type column, where ``Supervised'' denotes supervised methods, ``Self-supervised$\text{[S]}$'' represents single-modal self-supervised methods, and ``Self-supervised$\text{[C]}$'' refers to cross-modal self-supervised methods. All accuracy results are reported without employing the voting strategy.}
    \label{tab-PMT-MAE-cls}
\end{table}

Compared with supervised methods, PMT-MAE demonstrates competitive performance. For instance, PMT-MAE-L achieves an accuracy of 93.6\%, which is comparable to PointMLP~\cite{PointMLP2022}, the highest-performing supervised method listed, which achieves 94.1\%. This level of performance is noteworthy given that supervised methods benefit from labeled data during training, while PMT-MAE leverages an self-supervised approach during the pre-training phase. The competitive accuracy of PMT-MAE highlights its robustness and efficacy in extracting meaningful features from point clouds without the need for extensive labeled datasets.

When compared to cross-modal methods, PMT-MAE also shows comparable performance. For example, Joint-MAE~\cite{Joint-MAE2023}, a cross-modal method, achieves an accuracy of 94.0\%, while PMT-MAE-L reaches 93.6\%. Despite the slight edge in accuracy for Joint-MAE~\cite{Joint-MAE2023}, cross-modal methods have inherent disadvantages. They often require additional data from different modalities, which increases the complexity of data acquisition and preprocessing. Additionally, integrating multiple modalities can introduce significant computational overhead. In contrast, single-modal methods like PMT-MAE are simpler to implement and manage, as they do not require the integration of diverse data types. This simplicity can lead to more efficient and scalable solutions, particularly in resource-constrained environments.

Within the same category of single-modal self-supervised methods, both versions of PMT-MAE outperform the baseline model Point-MAE~\cite{Point-MAE2022} and the teacher model Point-M2AE~\cite{Point-M2AE2022}. Specifically, PMT-MAE-L surpasses the teacher model by achieving an accuracy of 93.6\% compared to 93.4\%. This improvement underscores the effectiveness of the proposed dual-branch architecture and the distillation strategies employed in PMT-MAE. The ability of PMT-MAE to outperform its predecessors highlights its potential for further advancements in single-modal self-supervised learning for point cloud classification.

Moreover, PMT-MAE requires only 40 epochs for both the pre-training and fine-tuning phases, demonstrating its efficiency. This efficiency is particularly beneficial for scenarios with limited computational resources, making PMT-MAE a practical and scalable solution for point cloud classification tasks. The reduced training time not only accelerates the development process but also allows for quicker iteration and optimization of the model, which is crucial in dynamic research and development environments.

PMT-MAE exhibits competitive performance on point cloud classification task, showcasing its robustness and efficiency compared to both supervised and self-supervised methods. Its single-modal approach offers simplicity and computational efficiency without compromising accuracy, positioning it as a promising model for practical point cloud analysis. The model's capacity to achieve high accuracy with minimal training epochs further underscores its potential for widespread application in the field of 3D point cloud processing.

\subsection{Ablation Study}

In this section, we conduct ablation experiments to evaluate the contributions of the dual-branch structure and the two-stage distillation process in our PMT-MAE models, illustrated in Tab.~\ref{tab-PMT-MAE-S-ablation} and Tab.~\ref{tab-PMT-MAE-L-ablation}. We compare four scenarios for both PMT-MAE-S and PMT-MAE-L, where the encoder consists of 6 and 12 blocks, respectively:
\begin{enumerate}
    \item Model-1: Trained from scratch with dual-branch block.
    \item Model-2: Distillation in the pre-training phase only (based on Model-1).
    \item Model-3: Two-stage distillation using the attention block of the traditional Transformer model (based on Model-2).
    \item PMT-MAE-S/PMT-MAE-L: Utilizes the dual-branch block and two-stage distillation.
\end{enumerate}

\begin{table}[ht]
    \centering
    \newcommand{\tabincell}[2]{\begin{tabular}{@{}#1@{}}#2\end{tabular}}
    \small
    \setlength{\tabcolsep}{1.0mm}
    \begin{tabular*}{\textwidth}{@{\extracolsep{\fill}}c|ccccc}
        \toprule[1pt]
        Model   
        & \tabincell{c}{Distil.\\(Pre-train)}       & \tabincell{c}{Distil.\\(Fine-tune)}   
        & \tabincell{c}{Attention}                  & \tabincell{c}{Dual-branch}    
        & Acc.(\%)\\
        \midrule[0.3pt]
        Model-1     &           &           &           & \ding{52} & 89.5  \\
        Model-2     & \ding{52} &           &           & \ding{52} & 92.3  \\
        Model-3     & \ding{52} & \ding{52} & \ding{52} &           & 92.6  \\
        \midrule[0.3pt]
        PMT-MAE-S   & \ding{52} & \ding{52} &           & \ding{52} & 93.5  \\
        \bottomrule[1pt]
    \end{tabular*}
    \vspace{5pt}
    \caption{Ablation study results for PMT-MAE-S. Note that all training involved in this table is limited to 40 epochs.}
    \label{tab-PMT-MAE-S-ablation}
\end{table}

\begin{table}[ht]
    \centering
    \newcommand{\tabincell}[2]{\begin{tabular}{@{}#1@{}}#2\end{tabular}}
    \small
    \setlength{\tabcolsep}{1.0mm}
    \begin{tabular*}{\textwidth}{@{\extracolsep{\fill}}c|ccccc}
        \toprule[1pt]
        Model   
        & \tabincell{c}{Distil.\\(Pre-train)}       & \tabincell{c}{Distil.\\(Fine-tune)}   
        & \tabincell{c}{Attention}                  & \tabincell{c}{Dual-branch}    
        & Acc.(\%)\\
        \midrule[0.3pt]
        Model-1     &           &           &           & \ding{52} & 84.7  \\
        Model-2     & \ding{52} &           &           & \ding{52} & 93.1  \\
        Model-3     & \ding{52} & \ding{52} & \ding{52} &           & 92.3  \\
        \midrule[0.3pt]
        PMT-MAE-L   & \ding{52} & \ding{52} &           & \ding{52} & 93.6  \\
        \bottomrule[1pt]
    \end{tabular*}
    \vspace{5pt}
    \caption{Ablation study results for PMT-MAE-L. Note that all training involved in this table is limited to 40 epochs.}
    \label{tab-PMT-MAE-L-ablation}
\end{table}

From the ablation results of PMT-MAE-S in Tab.~\ref{tab-PMT-MAE-S-ablation}, several insights can be drawn. Model-1 (Tab.~\ref{tab-PMT-MAE-S-ablation}), which is trained from scratch, achieves an accuracy of 89.5\%, indicating that without any distillation or sophisticated modules, the model's performance is significantly limited. In contrast, Model-2 (Tab.~\ref{tab-PMT-MAE-S-ablation}), which incorporates pre-training distillation, shows a marked improvement with an accuracy of 92.3\%. This highlights the critical role of the pre-training phase in boosting model performance. Model-3 (Tab.~\ref{tab-PMT-MAE-S-ablation}), which adds fine-tuning distillation with the traditional Transformer attention block, achieves a slightly higher accuracy of 92.6\%. Finally, PMT-MAE-S (Tab.~\ref{tab-PMT-MAE-S-ablation}), leveraging both the dual-branch structure and two-stage distillation, attains the highest accuracy of 93.5\%, underscoring the synergistic benefits of these components.

The ablation results for PMT-MAE-L (Tab.~\ref{tab-PMT-MAE-L-ablation}) further validate the effectiveness of the proposed architecture. Model-1 (Tab.~\ref{tab-PMT-MAE-L-ablation}), trained from scratch, yields an accuracy of 84.7\%, significantly lower than its counterpart in PMT-MAE-S. This suggests that deeper networks like PMT-MAE-L are more challenging to train effectively from scratch. Model-2 (Tab.~\ref{tab-PMT-MAE-L-ablation}), with pre-training distillation, shows a substantial improvement, achieving an accuracy of 93.1\%, highlighting the importance of the pre-training phase for deeper models. Model-3 (Tab.~\ref{tab-PMT-MAE-L-ablation}), incorporating fine-tuning distillation with the traditional Transformer attention block, achieves an accuracy of 92.3\%, lower than Model-2 (Tab.~\ref{tab-PMT-MAE-L-ablation}). This drop indicates that the traditional Transformer attention block may not be optimal for deeper architectures within limited epochs. PMT-MAE-L (Tab.~\ref{tab-PMT-MAE-L-ablation}), utilizing both the dual-branch structure and two-stage distillation, achieves the highest accuracy of 93.6\%, demonstrating the superior performance of the proposed approach.

When comparing the results in Tab.~\ref{tab-PMT-MAE-S-ablation} and Tab.~\ref{tab-PMT-MAE-L-ablation}, several common trends and anomalies become evident. A key observation is the superior performance of PMT-MAE-S over PMT-MAE-L for Model-1 and Model-3. Specifically, Model-1 (Tab.~\ref{tab-PMT-MAE-L-ablation}) for PMT-MAE-L shows a significantly lower accuracy (84.7\%) compared to Model-1 (Tab.~\ref{tab-PMT-MAE-S-ablation}) for PMT-MAE-S (89.5\%). Similarly, Model-3 (Tab.~\ref{tab-PMT-MAE-L-ablation}) for PMT-MAE-L achieves a lower accuracy (92.3\%) than Model-3 (Tab.~\ref{tab-PMT-MAE-S-ablation}) for PMT-MAE-S (92.6\%).

These discrepancies can be attributed to the depth of the network and the constraints imposed by the training epochs. For Model-1, which relies solely on the dual-branch module, the absence of a teacher model results in slower convergence and suboptimal performance for the deeper Model-1 (Tab.~\ref{tab-PMT-MAE-L-ablation}). For Model-3, which employs two-stage distillation but retains the traditional Transformer attention module, the deeper Model-3 (Tab.~\ref{tab-PMT-MAE-L-ablation}) also performs worse. This indicates that the traditional attention mechanism is less effective in deeper architectures under constrained training epochs, likely due to slower convergence rates and the inability to fully leverage the depth of the network within the limited training period.

Oppositely, compared to Tab.~\ref{tab-PMT-MAE-S-ablation}, the strong performance of Model-2 and PMT-MAE-L in Tab.~\ref{tab-PMT-MAE-L-ablation} underscores the decisive role of distillation during the pre-training phase for deeper models. This phase ensures faster convergence, allowing Model-2 and PMT-MAE-L in Tab.~\ref{tab-PMT-MAE-L-ablation} to fully exploit the enhanced representational capacity of their deeper network structures. Consequently, this leads to significant performance improvements, underscoring the importance of pre-training distillation in optimizing the performance of deeper networks.

Another anomaly is the performance trend from Model-2 to Model-3 in both tables. For Tab.~\ref{tab-PMT-MAE-S-ablation}, the accuracy increases from Model-2 (92.3\%) to Model-3 (92.6\%). In contrast, for Tab.~\ref{tab-PMT-MAE-L-ablation}, the accuracy decreases from Model-2 (93.1\%) to Model-3 (92.3\%). This divergence can be explained by the interaction between the network depth and the distillation process. In Tab.~\ref{tab-PMT-MAE-S-ablation}, the shallower architecture benefits from the additional fine-tuning distillation in Model-3, as the simpler structure converges more efficiently, allowing the model to make full use of the distillation process. However, in Tab.~\ref{tab-PMT-MAE-L-ablation}, while the fine-tuning distillation in Model-3 introduces beneficial guidance, the switch to the traditional Transformer attention module negatively impacts convergence speed. The deeper architecture's inherent slower convergence under the 40-epoch constraint exacerbates this issue, leading to a net decrease in performance from Model-2 to Model-3.

These findings highlight the critical role of both the dual-branch module and two-stage distillation in achieving fast convergence and high accuracy, particularly for deeper networks. The dual-branch structure provides rapid convergence, while two-stage distillation ensures comprehensive feature learning. The inefficacy of traditional attention mechanisms in deeper networks underscores the need for architectures and training strategies that can adapt to the depth of the model and the constraints of limited training epochs.

\subsection{Complexity Analysis}
This section presents a comparative analysis of the complexity and performance of the proposed PMT-MAE-S and PMT-MAE-L models against the baseline model Point-MAE~\cite{Point-MAE2022} and the teacher model Point-M2AE~\cite{Point-M2AE2022}. The key metrics for comparison include the number of parameters, floating point operations (FLOPs), the number of epochs for pre-training and fine-tuning, and accuracy without voting, as summarized in Tab.~\ref{PMT-MAE-tab-complexity}.

\begin{table}[ht]
    \centering
    \newcommand{\tabincell}[2]{\begin{tabular}{@{}#1@{}}#2\end{tabular}}
    \setlength{\tabcolsep}{0.0mm}
    \begin{tabular*}{\textwidth}{@{\extracolsep{\fill}}l|ccccc}
        \toprule[1pt]
        \textbf{Method}     & \tabincell{c}{Params.\\(M)}           & \tabincell{c}{FLOPs\\(G)}
                            & \tabincell{c}{Epochs\\(Pre-train)}    & \tabincell{c}{Epochs\\(Fine-tune)}
                            & \tabincell{c}{Acc.(\%)\\(w/o voting)} \\  
        \midrule[0.3pt]
        Point-MAE~\cite{Point-MAE2022}      & 22.1  & 2.4   & 300   & 300   & 93.2  \\
        Point-M2AE~\cite{Point-M2AE2022}    & 12.8  & 5.1   & 300   & 300   & 93.4  \\
        \midrule[0.3pt]
        PMT-MAE-S (Ours)                    & 14.0  & 1.9   & 40    & 40    & 93.5  \\
        PMT-MAE-L (Ours)                    & 27.3  & 2.7   & 40    & 40    & 93.6  \\
        \bottomrule[1pt]
    \end{tabular*}
    \vspace{5pt}
    \caption{Comparison of model complexity and performance between the proposed PMT-MAE-S and PMT-MAE-L models, the baseline model Point-MAE~\cite{Point-MAE2022}, and the teacher model Point-M2AE~\cite{Point-M2AE2022}, including the number of parameters, FLOPs, pre-training and fine-tuning epochs, and accuracy without voting.}
    \label{PMT-MAE-tab-complexity}
\end{table}

The proposed PMT-MAE-L model demonstrates a balanced trade-off between model complexity and performance. Compared to the baseline model Point-MAE~\cite{Point-MAE2022}, PMT-MAE-L incorporates an additional MLP branch in its dual-branch structure, resulting in a slight increase in both the number of parameters and FLOPs. Despite these increases, the performance gain, as indicated by the higher accuracy, justifies the added complexity. The lower number of epochs (40 for both pre-training and fine-tuning) underscores the efficiency of the PMT-MAE-L model in achieving convergence faster than the baseline, which requires 300 epochs for each phase.

When compared to the teacher model Point-M2AE~\cite{Point-M2AE2022}, PMT-MAE-L maintains a higher number of parameters due to its deeper architecture. However, the FLOPs for PMT-MAE-L are substantially lower than those for Point-M2AE. This discrepancy is attributed to the pyramid structure of Point-M2AE, which, although reducing the number of parameters, increases the number of tokens processed at the shallow stages, thereby elevating the FLOPs. Conversely, PMT-MAE-L, with fewer tokens processed, achieves a lower computational burden while still leveraging the stronger characterization capability of its deeper network to attain superior performance.

The PMT-MAE-S model, a shallower variant, exhibits the lowest FLOPs among all models due to its reduced number of encoder modules (6 blocks). This reduction significantly decreases both the parameters and computational complexity while maintaining a competitive accuracy of 93.5\%, which is higher than both the baseline Point-MAE~\cite{Point-MAE2022} and the teacher Point-M2AE~\cite{Point-M2AE2022}. The efficiency of PMT-MAE-S, evident from the lower FLOPs and faster convergence within 40 epochs, highlights the advantage of its streamlined architecture without compromising on performance.

In summary, the PMT-MAE models, both S and L variants, effectively balance model complexity and performance. The dual-branch structure and two-stage distillation strategy enable these models to achieve high accuracy with significantly fewer epochs compared to traditional approaches. The results validate the efficacy of the proposed PMT-MAE architecture in point cloud classification tasks, demonstrating its potential for future advancements in the field.

\subsection{Effect of Pre-training Distillation Weight \(\alpha\)}
In the pre-training phase, feature distillation between the teacher and student models is controlled by the distillation weight, denoted as \(\alpha\). This subsection examines how different values of \(\alpha\) influence the classification performance of the proposed PMT-MAE-S and PMT-MAE-L models. The results of these experiments are summarized in Tab.~\ref{PMT-MAE-tab-alpha}.

\begin{table}[ht]
    \centering
    \newcommand{\tabincell}[2]{\begin{tabular}{@{}#1@{}}#2\end{tabular}}
    \small
    \setlength{\tabcolsep}{4.0mm}
    \begin{tabular*}{\textwidth}{@{\extracolsep{\fill}}c|ccccccc}
        \toprule[1pt]
        $\alpha$                & 0.1   & 0.5   & 1.0   & 2.0   & 3.0   & 4.0   & 5.0   \\
        \midrule[0.3pt]
        Acc. (S)                & 92.9  & 93.3  & 93.5  & 93.3  & 93.1  & 93.1  & 92.9  \\
        Acc. (L)                & 93.1  & 93.3  & 93.6  & 93.4  & 93.4  & 93.1  & 93.0  \\
        \bottomrule[1pt]
    \end{tabular*}
    \vspace{5pt}
    \caption{Effect of pre-training distillation weight \(\alpha\) on the classification accuracy of PMT-MAE-\textbf{S} and PMT-MAE-\textbf{L} models (\%).}
    \label{PMT-MAE-tab-alpha}
\end{table}

The experimental setup involved varying \(\alpha\) from 0.1 to 5.0 and measuring the classification accuracy of PMT-MAE-S and PMT-MAE-L. The results indicate that the optimal value of \(\alpha\) for both models is 1.0, where PMT-MAE-S achieves an accuracy of 93.5\% and PMT-MAE-L reaches 93.6\%. Lower values of \(\alpha\) result in reduced accuracy, while higher values also lead to a decline in performance.

A deeper analysis reveals that an \(\alpha\) value of 1.0 provides the best balance between the influence of the teacher model and the adaptability of the student model. When \(\alpha\) is too low, the student model does not receive sufficient guidance from the teacher, leading to suboptimal learning outcomes. Conversely, excessively high \(\alpha\) values cause the student model to rely too heavily on the teacher, which can hinder its ability to learn independently and generalize from the data.

These findings underscore the importance of tuning the distillation weight \(\alpha\) during the pre-training phase to optimize model performance. Achieving the right balance allows the student model to benefit from the teacher's knowledge while maintaining its own learning flexibility.

\subsection{Effect of Fine-tuning Distillation Weight \(\beta\) and Temperature \(T\)}
This subsection investigates the impact of the distillation weight \(\beta\) and the temperature \(T\) on the performance of the models during the fine-tuning stage of distillation. The parameters \(\beta\) and \(T\) were systematically varied to observe their effects on the classification accuracy of the PMT-MAE-S and PMT-MAE-L models.

To evaluate the effect of the distillation weight \(\beta\), we fixed the temperature \(T\) at 3.0 and varied \(\beta\) across several values. As shown in Tab.~\ref{PMT-MAE-tab-beta}, the best performance for both PMT-MAE-S and PMT-MAE-L was achieved at \(\beta = 0.01\), with PMT-MAE-S reaching an accuracy of 93.5\% and PMT-MAE-L reaching 93.6\%. Increasing or decreasing \(\beta\) from this value resulted in a noticeable decline in accuracy, indicating the sensitivity of model performance to the distillation weight during fine-tuning.

\begin{table}[ht]
    \centering
    \newcommand{\tabincell}[2]{\begin{tabular}{@{}#1@{}}#2\end{tabular}}
    \small
    \setlength{\tabcolsep}{8.0mm}
    \begin{tabular*}{\textwidth}{@{\extracolsep{\fill}}c|cccc}
        \toprule[1pt]
        $\beta$                 & 0.001 & 0.01  & 0.1   & 1.0   \\
        \midrule[0.3pt]
        Acc. (S)                & 92.9  & 93.5  & 93.3  & 92.7  \\
        Acc. (L)                & 93.5  & 93.6  & 92.6  & 92.2  \\
        \bottomrule[1pt]
    \end{tabular*}
    \vspace{5pt}
    \caption{Effect of fine-tuning distillation weight \(\beta\) on performance (\%). ``S'' and ``L'' correspond to the PMT-MAE-S and PMT-MAE-L models.}
    \label{PMT-MAE-tab-beta}
\end{table}

Similarly, we investigated the effect of the temperature \(T\) by fixing \(\beta\) at 0.01 and varying \(T\) across a range of values. The results, presented in Tab.~\ref{PMT-MAE-tab-T}, show that the optimal temperature for both PMT-MAE-S and PMT-MAE-L is \(T = 3.0\), where the models achieved their highest accuracy, 93.5\% and 93.6\% respectively. Variations in \(T\) either above or below this value led to reduced performance, emphasizing the critical role of temperature in the distillation process during fine-tuning.

\begin{table}[ht]
    \centering
    \newcommand{\tabincell}[2]{\begin{tabular}{@{}#1@{}}#2\end{tabular}}
    \small
    \setlength{\tabcolsep}{6.0mm}
    \begin{tabular*}{\textwidth}{@{\extracolsep{\fill}}c|ccccc}
        \toprule[1pt]
        $T$                     & 1.0   & 2.0   & 3.0   & 4.0   & 5.0   \\
        \midrule[0.3pt]
        Acc. (S)                & 93.0  & 93.0  & 93.5  & 93.1  & 92.9  \\
        Acc. (L)                & 93.4  & 93.6  & 93.6  & 93.2  & 93.1  \\
        \bottomrule[1pt]
    \end{tabular*}
    \vspace{5pt}
    \caption{Effect of fine-tuning distillation temperature \(T\) on performance (\%). The symbols ``S'' and ``L'' denote PMT-MAE-S and PMT-MAE-L models.}
    \label{PMT-MAE-tab-T}
\end{table}

In summary, the experiments demonstrate that both the distillation weight \(\beta\) and the temperature \(T\) significantly influence the fine-tuning performance of PMT-MAE models. The optimal settings identified were \(\beta = 0.01\) and \(T = 3.0\), which provided the highest classification accuracy for both PMT-MAE-S and PMT-MAE-L. These findings highlight the importance of tuning these parameters to achieve optimal performance during the fine-tuning stage of distillation.

\subsection{Effect of Mask Ratio on Performance}
This subsection examines the influence of the mask ratio during the pre-training stage on model performance. The mask ratio, which dictates the proportion of input data that is masked and subsequently reconstructed by the model, plays a crucial role in self-supervised learning frameworks. To investigate its effect, we conducted a series of experiments with varying mask ratios and evaluated the classification accuracy of the PMT-MAE-S and PMT-MAE-L models.

The experimental results are summarized in Tab.~\ref{PMT-MAE-tab-mask-ratio}. For both PMT-MAE-S and PMT-MAE-L, we tested mask ratios of 0.4, 0.5, 0.6, 0.7, and 0.8. We selected this range because it encompasses a moderate span of mask ratios, avoiding the extremes of very low or very high values. This range allows us to observe the model's behavior under different masking intensities without skewing the results due to excessive masking or insufficient masking.

\begin{table}[ht]
    \centering
    \newcommand{\tabincell}[2]{\begin{tabular}{@{}#1@{}}#2\end{tabular}}
    \small
    \setlength{\tabcolsep}{7.0mm}
    \begin{tabular*}{\textwidth}{@{\extracolsep{\fill}}c|ccccc}
        \toprule[1pt]
        Mask-ratio  & 0.4   & 0.5   & 0.6   & 0.7   & 0.8   \\
        \midrule[0.3pt]
        Acc. (S)    & 92.9  & 93.1  & 92.9  & 93.5  & 92.9  \\
        Acc. (L)    & 92.8  & 93.0  & 93.1  & 93.6  & 92.8  \\
        \bottomrule[1pt]
    \end{tabular*}
    \vspace{5pt}
    \caption{Classification results for various mask ratios (\%). ``S'' and ``L'' indicate the PMT-MAE-S and PMT-MAE-L models, respectively.}
    \label{PMT-MAE-tab-mask-ratio}
\end{table}

The results indicate that the highest accuracy for PMT-MAE-S, 93.5\%, was achieved with a mask ratio of 0.7. Other mask ratios yielded slightly lower accuracies, indicating that the model benefits from a balanced approach to masking. Similarly, PMT-MAE-L showed the best performance with a mask ratio of 0.7, achieving an accuracy of 93.6\%. This suggests that both model sizes favor a moderately high mask ratio during pre-training.

The choice of mask ratio range (0.4 to 0.8) was intentional to ensure that the model is sufficiently challenged to reconstruct the masked inputs while not being overwhelmed by excessive masking. Very low mask ratios may not provide enough difficulty for the model to develop strong reconstruction capabilities, whereas very high mask ratios may obscure too much information, making the learning task excessively difficult. The optimal mask ratio of 0.7 strikes a balance, allowing the model to effectively learn useful representations from the masked data.

In summary, the experiments indicate that the mask ratio is a critical hyperparameter in the pre-training phase of PMT-MAE models. A mask ratio of 0.7 was found to be optimal for both PMT-MAE-S and PMT-MAE-L, leading to the highest classification accuracies.

\subsection{Correlation Analysis in Double-Branch Structure}
In this study, a parallel dual-branch structure comprising Transformer and MLP is proposed as the basic module of the model. To verify that the output features of the two branches are weakly correlated and that the dual-branch setup enhances feature diversity, an analysis was conducted using the PMT-MAE-L and PMT-MAE-S models. The PMT-MAE-L encoder consists of 12 blocks, while the PMT-MAE-S encoder has 6 blocks. For each block, 2,468 samples from the test set were computed, with each sample corresponding to 64 tokens, resulting in a total of 157,952 tokens processed through the dual-branch structure of each block.

\begin{figure}[htbp]
    \centering
    \includegraphics[width=1.0\linewidth]{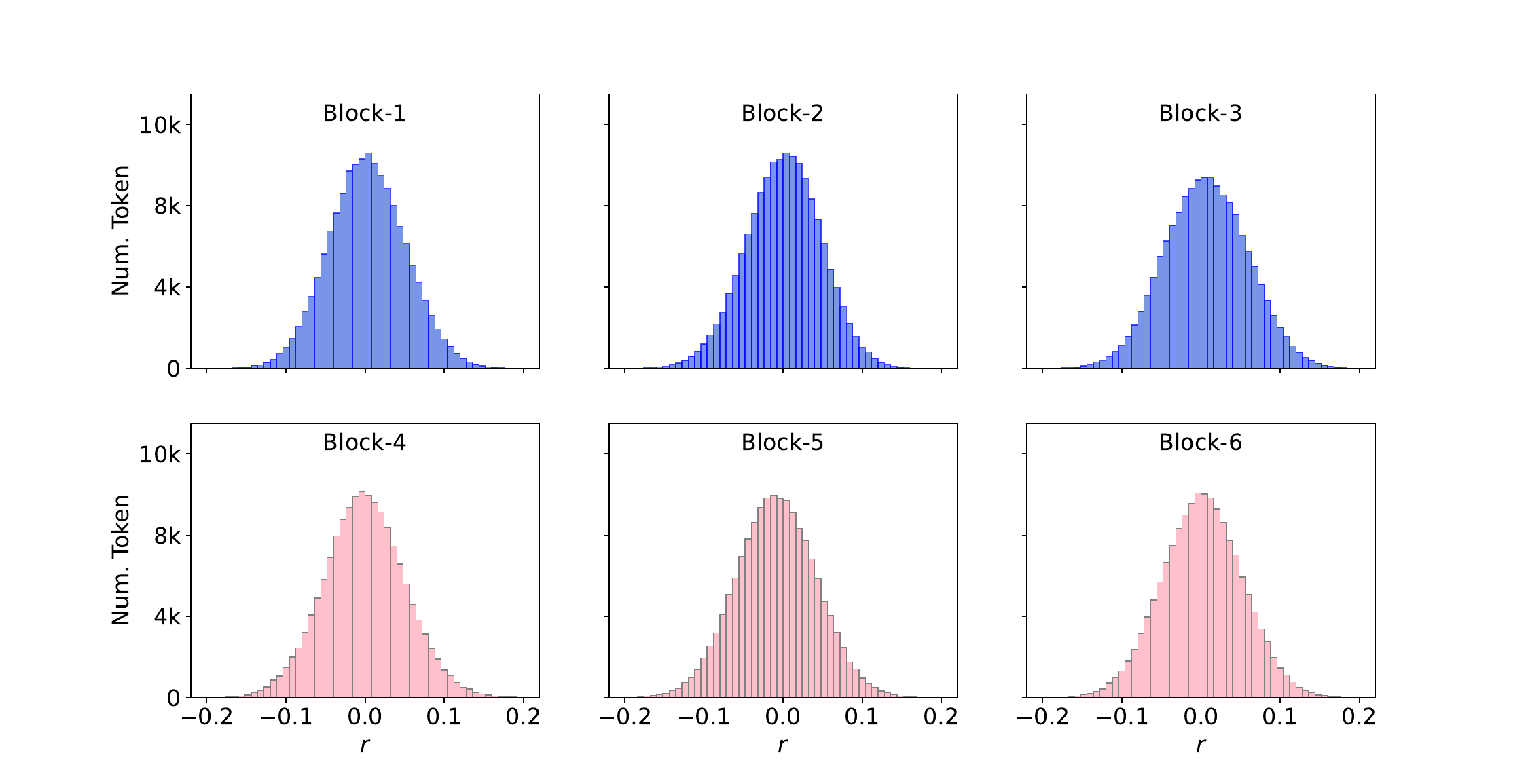}
    \caption{Histogram of the Pearson correlation coefficient (\(r\)) distribution for the output features of the two branches across six blocks in the PMT-MAE-S model. Each block's distribution is shown as a separate subdiagram.}
    \label{fig-PMT-MAE-S-pearson}
\end{figure}

\begin{figure}[htbp]
    \centering
    \includegraphics[width=1.0\linewidth]{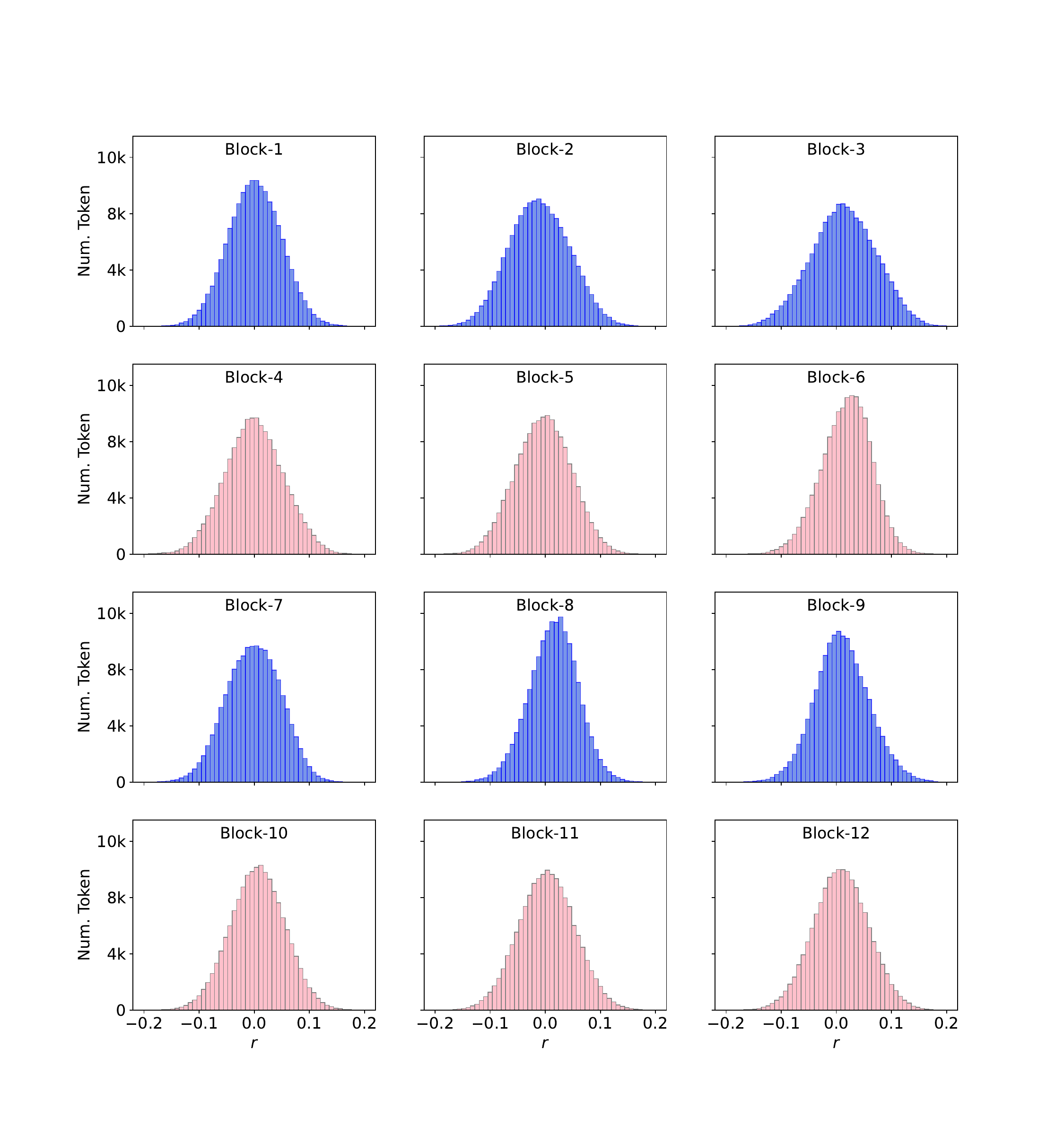}
    \caption{Histogram of the Pearson correlation coefficient (\(r\)) distribution for the output features of the two branches across twelve blocks in the PMT-MAE-L model. Each block's distribution is shown as a separate subdiagram.}
    \label{fig-PMT-MAE-L-pearson}
\end{figure}

The Pearson correlation coefficient \(r\) of the output features from the two branches was calculated for these tokens, and a histogram representing the statistical distribution of the number of tokens relative to \(r\) was plotted. This resulted in a histogram of the statistical distribution for 6 blocks in PMT-MAE-S (see Fig. ~\ref{fig-PMT-MAE-S-pearson}) and a histogram for 12 blocks in PMT-MAE-L (see Fig. ~\ref{fig-PMT-MAE-L-pearson}).

From these plots, it can be observed that the overall distribution of the number of tokens with respect to \(r\) resembles a normal distribution, centered around a mean of 0 and spread within a range of ±0.2. This indicates that the linear correlation between the two branches is minimal, suggesting that the features generated by each branch capture different aspects of the point cloud data. Consequently, the two branches complement each other, enhancing the diversity and characterization of the features.

\subsection{t-SNE Visualization of Encoder Features}
\begin{figure}[htbp]
    \centering
    \includegraphics[width=1.0\linewidth]{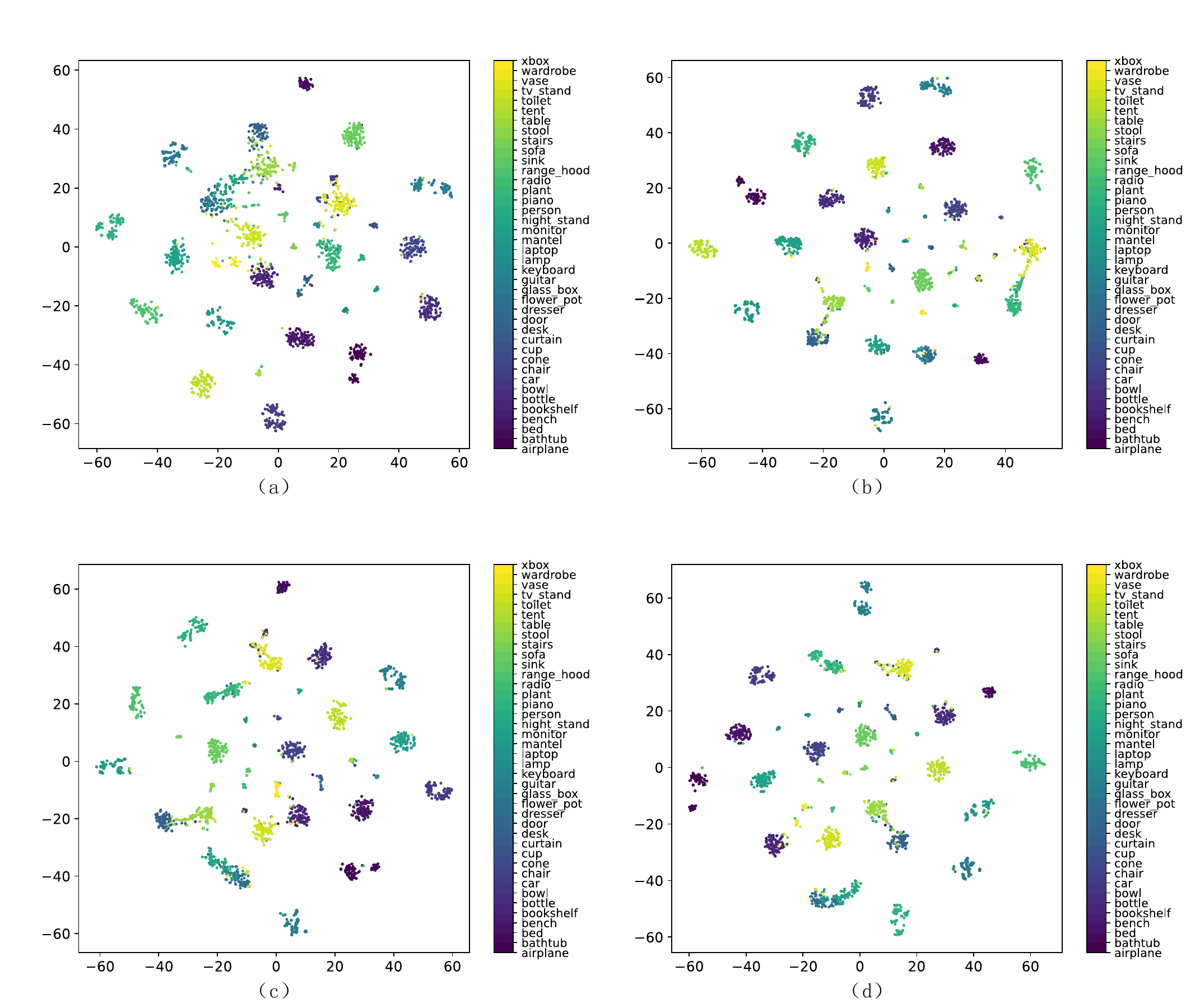}
    \caption{t-SNE visualization of the encoder features of the test set for four models. (a) Point-MAE, (b) Point-M2AE, (c) PMT-MAE-S, and (d) PMT-MAE-L. The clear separability in all models is attributed to the fusion of the class-token and pooled features, which together form the final encoded features and enhance the models' ability to differentiate between classes.}
    \label{fig-PMT-MAE-tSNE}
\end{figure}

In this study, we employed t-distributed Stochastic Neighbor Embedding (t-SNE) visualization to analyze the encoder features of the test set for four models: the baseline model Point-MAE~\cite{Point-MAE2022}, the teacher model Point-M2AE~\cite{Point-M2AE2022}, and the proposed PMT-MAE-S and PMT-MAE-L models. All four models incorporate a class-token that interacts attentively with other tokens during the encoding process. This class-token, combined with pooled features, leads to highly separable encoded features, enhancing the ability to distinguish between different classes. The visualization results are presented in Fig.~\ref{fig-PMT-MAE-tSNE}, which includes four subplots, denoted as (a), (b), (c), and (d).

In Fig.~\ref{fig-PMT-MAE-tSNE}(a), which corresponds to Point-MAE~\cite{Point-MAE2022}, we observe that the boundaries of clusters in certain regions are indistinct, and there is significant overlap between categories. This indicates that the feature discrimination ability of Point-MAE~\cite{Point-MAE2022} is relatively weak. The clusters lack clear separation, suggesting that the model struggles to distinguish between different classes effectively.

In contrast, Fig.~\ref{fig-PMT-MAE-tSNE}(b) shows the visualization for Point-M2AE~\cite{Point-M2AE2022}. Compared to Point-MAE~\cite{Point-MAE2022}, the internal distribution of individual clusters is more concentrated, and the intervals between clusters of different categories are more distinct. This demonstrates that Point-M2AE~\cite{Point-M2AE2022} achieves better separability, enhancing its ability to differentiate between classes.

Fig.~\ref{fig-PMT-MAE-tSNE}(c) illustrates the results for PMT-MAE-S. Overall, the distribution is similar to that of Point-M2AE~\cite{Point-M2AE2022}. However, for clusters with a large number of samples, the concentration is not as high as in Fig.~\ref{fig-PMT-MAE-tSNE}(b). Additionally, the clusters with a smaller number of samples are less discrete compared to Point-M2AE~\cite{Point-M2AE2022}. Despite these differences, under the guidance of the teacher model Point-M2AE~\cite{Point-M2AE2022}, PMT-MAE-S maintains significant distinguishability overall.

Finally, Fig.~\ref{fig-PMT-MAE-tSNE}(d) displays the visualization for PMT-MAE-L. The distribution patterns are almost identical to those in Fig.~\ref{fig-PMT-MAE-tSNE}(b), suggesting that PMT-MAE-L effectively mimics the behavior of Point-M2AE~\cite{Point-M2AE2022}. This indicates that the dual-branch structure of PMT-MAE-L successfully captures and replicates the discriminative features learned by Point-M2AE~\cite{Point-M2AE2022}, further confirming the effectiveness of the proposed architecture.

Overall, the t-SNE visualizations provide valuable insights into the feature representation capabilities of the four models. The proposed PMT-MAE-L show clear advantages in feature separability and concentration compared to the baseline Point-MAE~\cite{Point-MAE2022}, highlighting the benefits of the two-stage distillation framework and the dual-branch architecture. The results for PMT-MAE-S, while slightly less concentrated for large sample clusters, still demonstrate significant improvements in feature discrimination over Point-MAE~\cite{Point-MAE2022}. These findings support the effectiveness of the proposed models in enhancing feature diversity and classification performance, thereby validating our approach.

\subsection{Visual Reconstruction of Point Clouds}
\begin{figure}[tbp]
    \centering
    \includegraphics[width=0.95\linewidth]{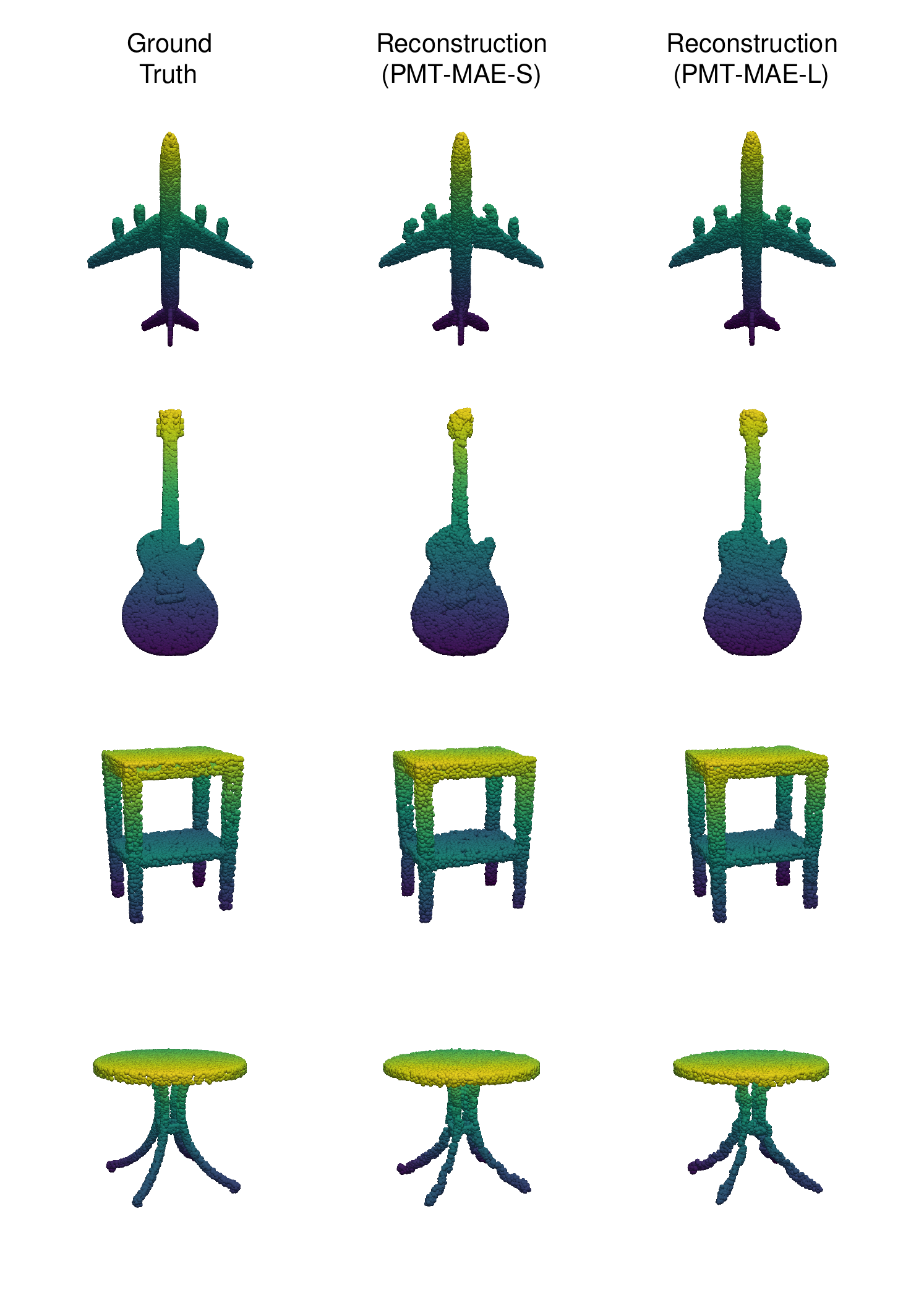}
    \caption{Visual comparison of point cloud reconstruction by PMT-MAE models against ground truth. Each row corresponds to a different sample, with columns representing (from left to right) Ground Truth, Reconstruction by PMT-MAE-S, and Reconstruction by PMT-MAE-L. The pre-training model's masking ratio was set to 0.7.}
    \label{fig-PMT-MAE-Reconstruction}
\end{figure}

This subsection investigates the masking and reconstruction capabilities of the pre-trained PMT-MAE models on point cloud. To assess reconstruction quality, we visualized the model's output compared to the original input point cloud, with results displayed in Fig.~\ref{fig-PMT-MAE-Reconstruction}. The figure includes four sample rows and three columns representing Ground Truth, Reconstruction (PMT-MAE-S), and Reconstruction (PMT-MAE-L). A mask ratio of 0.7 was applied during the pre-training phase.

The reconstruction results demonstrate the models' proficiency in retaining the overall shape of the original point clouds, even in regions where the point density is sparse. The PMT-MAE models effectively fill in the voids, suggesting a robustness in the self-supervised learning approach. However, upon closer inspection, it is observed that certain localized fine structures and elongated shapes exhibit suboptimal reconstruction. This indicates areas where the model's performance could be enhanced, potentially through architectural improvements or adjustments in the pre-training regime.

\section{Conclusion}
In this study, we employed a parallel dual-branch structure comprising Transformer and MLP modules, integrated into the PMT-MAE models for point cloud analysis, along with a two-stage distillation strategy. Our comprehensive experiments demonstrated that the PMT-MAE models outperform the baseline Point-MAE~\cite{Point-MAE2022} and the teacher model Point-M2AE~\cite{Point-M2AE2022} in various aspects, including computational efficiency and feature diversity. Specifically, the PMT-MAE models exhibit a lower FLOPs and higher classification accuracy, confirming their effectiveness and efficiency. Future research should focus on further enhancing pre-training techniques and exploring advanced reconstruction strategies to improve the model's ability to handle complex structures. Additionally, due to the limitation of computing resources, this study was not able to apply the model to more complex point cloud data. Future research can extend the application to other point cloud analysis tasks to further validate the utility of the proposed model.

\bibliographystyle{elsarticle-num}
\bibliography{reference}
\end{document}